\documentclass[runningheads]{llncs}

\usepackage[T1]{fontenc}
\usepackage{graphicx}     
\usepackage{amsmath,amsfonts}         
\usepackage{booktabs,tabularx}
\usepackage[table]{xcolor}            
\usepackage{hyperref}                 
\usepackage{makecell}

\DeclareUnicodeCharacter{2009}{\,}    
\newcolumntype{C}{>{\centering\arraybackslash}X} 

\title{Progressive Weight Loading: Accelerating Initial Inference and Gradually Boosting Performance on Resource‑Constrained Environments}
\titlerunning{Progressive Weight Loading}

\author{Hyunwoo Kim\inst{1} \and
        Junha Lee\inst{2} \and
        Mincheol Choi\inst{2} \and
        Jeonghwan Lee\inst{2} \and 
        Jaeshin Cho\inst{2} 
        }
\authorrunning{H. Kim et al.}

\institute{Intel, Seoul, South Korea\\
           \email{onion.kim@intel.com}
           \and
           Yonsei University, Seoul, South Korea\\
           \email{\{junha4304,charmfe,hwan1997,cjs823\}@yonsei.ac.kr}}

\begin{document}
\maketitle

\begin{abstract}
  Deep learning models have become increasingly large and complex, resulting in
  higher memory consumption and computational demands. Consequently, model
  loading times and initial inference latency have increased, posing significant
  challenges in mobile and latency-sensitive environments where frequent model
  loading and unloading are required, which directly impacts user experience.
  While Knowledge Distillation (KD) offers a solution by compressing large
  teacher models into smaller student ones, it often comes at the cost of reduced
  performance. To address this trade-off, we propose Progressive Weight Loading
  (PWL), a novel technique that enables fast initial inference by first deploying
  a lightweight student model, then incrementally replacing its layers with
  those of a pre-trained teacher model. To support seamless layer substitution,
  we introduce a training method that not only aligns intermediate feature
  representations between student and teacher layers, but also improves the
  overall output performance of the student model. Our experiments on VGG,
  ResNet, and ViT architectures demonstrate that models trained with PWL maintain
  competitive distillation performance and gradually improve accuracy as teacher
  layers are loaded—matching the final accuracy of the full teacher model without
  compromising initial inference speed. This makes PWL particularly suited for
  dynamic, resource-constrained deployments where both responsiveness and
  performance are critical.

\end{abstract}

\section{Introduction}

With the rapid advancement of deep learning, modern AI models have grown
substantially in both size and
complexity\cite{DBLP:journals/corr/abs-2001-08361}, leading to increased
computational and memory
demands\cite{DBLP:journals/corr/abs-1906-02243,DBLP:journals/corr/abs-2104-10350}.
While larger models often deliver improved performance, their size also results
in longer loading times, which can become a critical bottleneck during
deployment\cite{han2016deepcompressioncompressingdeep,DBLP:journals/corr/HowardZCKWWAA17}.
In memory-constrained environments like edge devices, these loading delays can
be especially severe. A recent study reported that for multi-billion-parameter
models, the time spent loading the model into GPU memory can account for over
90\% of the total inference latency\cite{Han2024Hermes}. Fast and efficient
model loading is especially important in latency sensitive applications such as
autonomous driving, robotics, and interactive AI services, where delays in
model initialization may adversely affect user experience or even compromise
safety.

One of the simplest approaches to reducing model loading time is to make the
model smaller. Knowledge distillation is a widely used technique that
compresses large models into smaller ones while attempting to retain their
performance\cite{hinton2015distilling,buciluǎ2006model}. In this process,
student models are trained under the guidance of larger teacher models by
mimicking the teacher's output logits\cite{hinton2015distilling} and
intermediate features\cite{romero2014fitnets}. However, the performance of
student models often falls short of that of their teacher
models\cite{gou2021knowledge} and is generally constrained by the student
model's size. This means that although smaller student models can achieve
faster loading times, they may struggle to match the accuracy of their larger
counterparts.

\begin{figure}[htbp]

  \centering

  \includegraphics[width=1.0\linewidth]{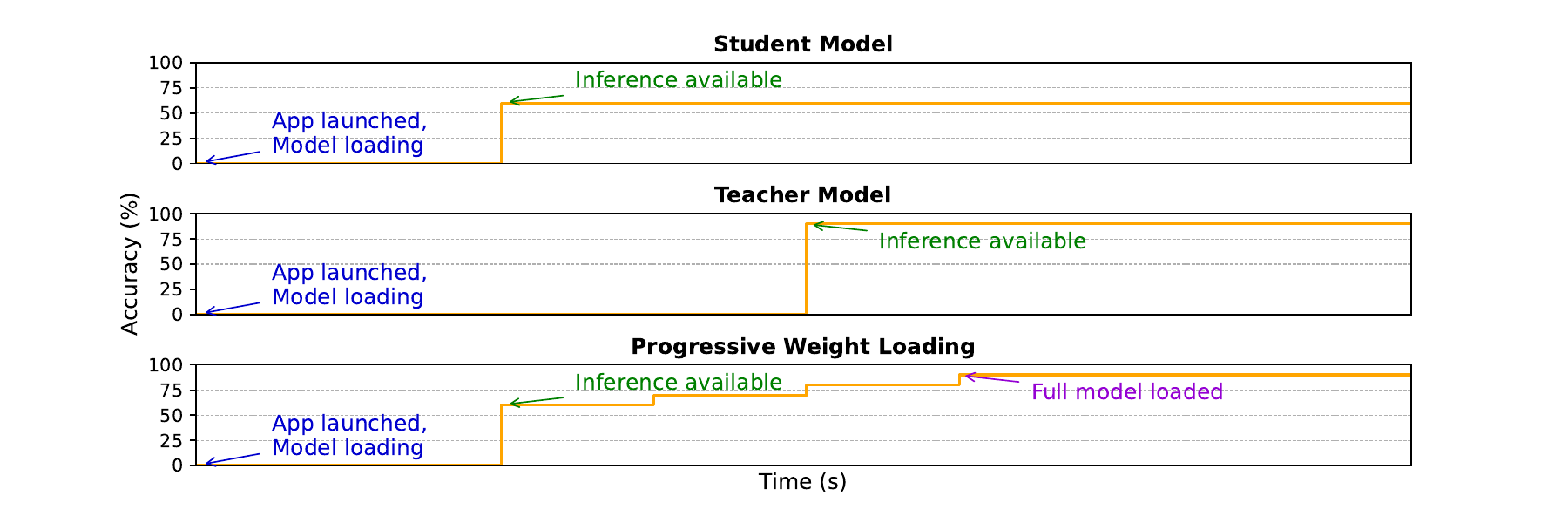}

  \caption{Illustration of the Progressive Weight Loading (PWL) timeline. Initial inference is performed using the lightweight student model, enabling a fast response. As layers are progressively replaced with those of the teacher model, the overall performance improves gradually toward that of the full teacher.}

  \label{fig:pwl_timeline}

\end{figure}

In this paper, we propose a novel approach called Progressive Weight Loading
(PWL) to tackle the challenge of balancing model loading time and performance.
As illustrated in Figure~\ref{fig:pwl_timeline}, PWL achieves fast loading by
initially loading a smaller student model, then progressively loading the
teacher model's weights layer by layer while enhancing the student's
performance. This strategy allows us to strike a balance between rapid loading
and high accuracy, making it particularly suitable for resource-constrained
mobile and latency-sensitive environments. Additionally, we introduce a
distillation training method and a student model architecture designed to
enable seamless performance improvement as layers of the student model are
incrementally replaced by those of the teacher model.

\section{Background}

Deploying deep neural networks on resource-constrained edge devices—such as
smartphones, IoT systems, and AR/VR headsets—poses significant challenges due
to limited compute, memory, and power resources compared to cloud
environments\cite{chen2019deep,shuvo2022efficient}. These constraints often
necessitate model loading and unloading at runtime, making techniques that
accelerate model loading and enable fast initial inference highly valuable.

To address these challenges, various model compression techniques have been
developed. Among them,
pruning\cite{han2016deepcompressioncompressingdeep,molchanov2016pruning,molchanov2017variational}
and quantization\cite{hubara2018quantized,frantar2022gptq} reduce model
size and computational cost by removing redundant parameters or lowering
numerical precision. Another widely studied method is Knowledge Distillation
(KD)\cite{hinton2015distilling}, where a compact student model is trained to
mimic the behavior of a larger, high-performing teacher model. By learning from
the teacher's outputs, the student can achieve competitive accuracy while
maintaining a smaller footprint.

KD has proven effective in real world applications, with models like
MobileNet\cite{DBLP:journals/corr/HowardZCKWWAA17,sandler2018mobilenetv2,howard2019searching},
TinyBERT\cite{jiao2019tinybert}, EfficientNet\cite{tan2019efficientnet}, and
MobileBERT\cite{sun2020mobilebert} demonstrating its impact in enabling
efficient deep learning on edge devices. Initially, KD focused on matching the
final outputs (logits) of the teacher and student
networks\cite{hinton2015distilling}. However, recent research has extended this
concept to hinted loss\cite{romero2014fitnets}, which includes an intermediate
feature alignment step. In this step, the student is guided to match the
teacher's internal representations, leading to more robust and informative
knowledge transfer. Methods such as layer-wise
distillation\cite{sun2020mobilebert,tung2019similarity} and attention
transfer\cite{zagoruyko2016paying,ji2021show} exemplify this progression.
Attention transfer proposes re-weighting feature maps with attention maps for
loss construction\cite{zagoruyko2016paying}. ReviewKD\cite{chen2021distilling}
addresses the challenge of arbitrarily selecting specific layers for
distillation by aggregating information across all teacher layers using
trainable attention blocks. Neuron Selectivity Transfer\cite{huang2017like} and
Similarity Preserving KD\cite{tung2019similarity} involve the construction of
relational matrices based on batches and features; these matrices then serve as
input for their distillation loss functions.

Despite these advancements, the primary objective of existing intermediate
feature alignment techniques is to improve the standalone performance of the
student model\cite{gou2021knowledge}. In contrast, our work introduces a novel
perspective: training a student model whose layers can be seamlessly replaced
by corresponding layers from the teacher model at inference time, without
degrading performance. This approach enables Progressive Weight Loading (PWL)—a
mechanism that supports fast initial inference and gradually upgrades the model
toward full teacher level performance in a resource aware manner.

\section{Design Proposal}

\subsection{Progressive Weight Loading Overview}

\begin{figure}[htbp]

  \centering

  \includegraphics[width=1.0\linewidth]{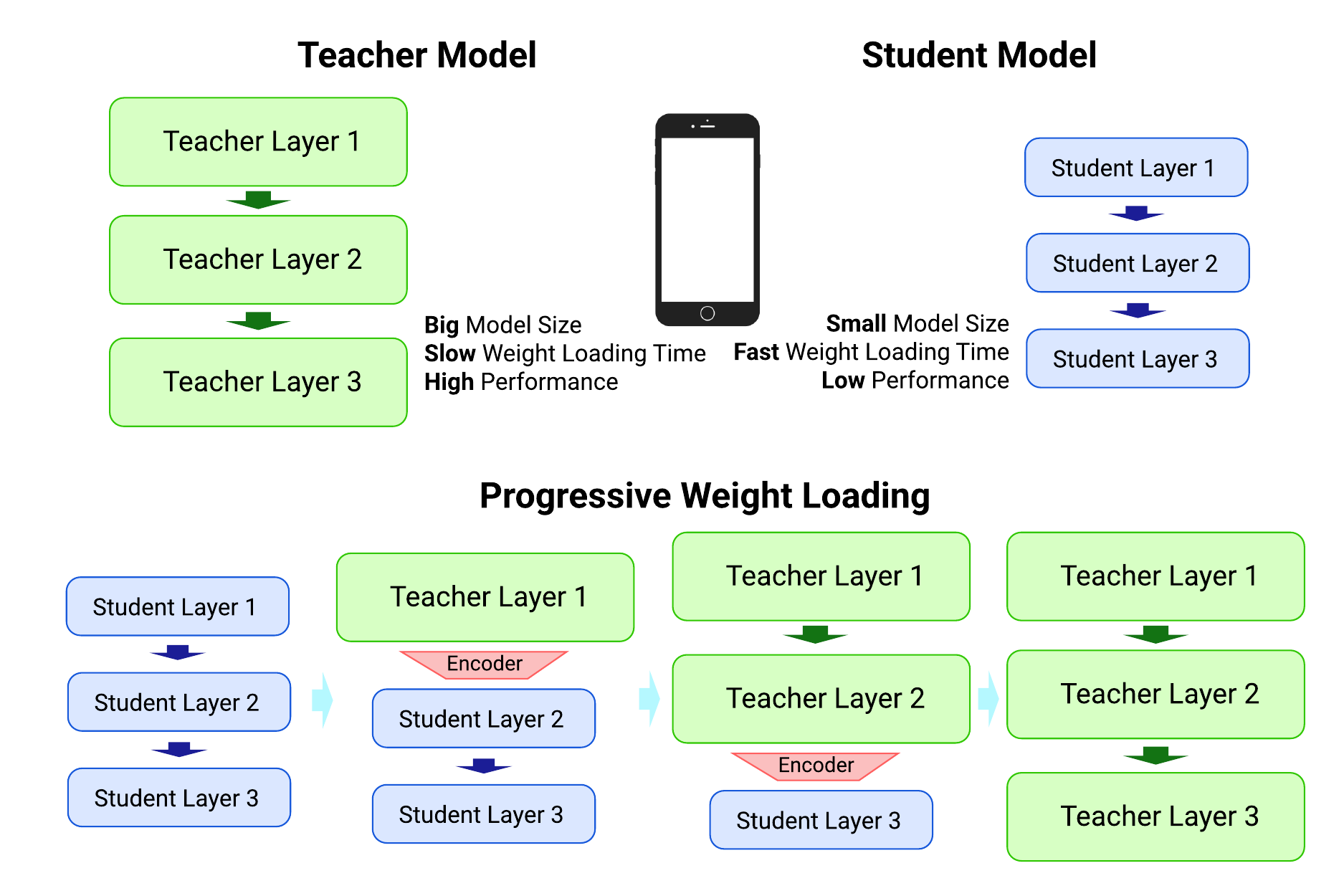}

  \caption{Conceptual overview of Progressive Weight Loading (PWL). The student model is initially loaded, and its layers are progressively replaced by those of the teacher model, starting from the input layer. This approach enables a dynamic trade-off between model size and performance, making it well-suited to resource-constrained environments such as mobile and edge computing.}

  \label{fig:pwl_concept}

\end{figure}

PWL starts inference with a smaller, efficient student model and incrementally
replaces its layers with those from a larger, more powerful teacher model. The
objective of PWL is to progressively enhance performance from the baseline
level of the student to the superior level of the teacher. Key benefits of PWL
include:

\begin{enumerate}
  \item \textbf{Fast Initial Inference}: By initially deploying the smaller student model,
        PWL enables rapid model loading and inference, critical for latency-sensitive applications
        such as autonomous driving or interactive AI services.
  \item \textbf{Memory Aware Performance Scaling}: PWL provides dynamic control over the
        memory performance trade-off, making it particularly suitable for mobile and resource-constrained
        environments. Unlike conventional deployments—which rely entirely on either a small student or a large
        teacher model, PWL supports partial loading of teacher layers according to the available system memory.
\end{enumerate}

The primary goal of PWL is to achieve gradual and seamless performance
improvement as student model's layers are incrementally replaced with
corresponding layers from the teacher model. However, this approach poses
several challenges:

\begin{itemize}
  \item \textbf{Feature Alignment}: Differences in architecture between student
        and teacher models complicate the alignment of their internal feature representations.
        Therefore, a dedicated feature converter is required to map the student's features to the teacher's
        feature space and vice versa.

  \item \textbf{Training Strategy}: An effective training strategy must be developed to prevent
        performance degradation during progressive layer replacement. This strategy should facilitate
        the seamless alignment of student and teacher features, ensuring the student model effectively
        mimics both the logits and the intermediate representations of the teacher.
\end{itemize}

To address these challenges, we introduce two key components: the
\textbf{Invertible Feature Converter} and the \textbf{Training Strategy for
  PWL}.

\subsection{Invertible Feature Converter}

\begin{figure}[htbp]

  \centering

  \includegraphics[width=1.0\linewidth]{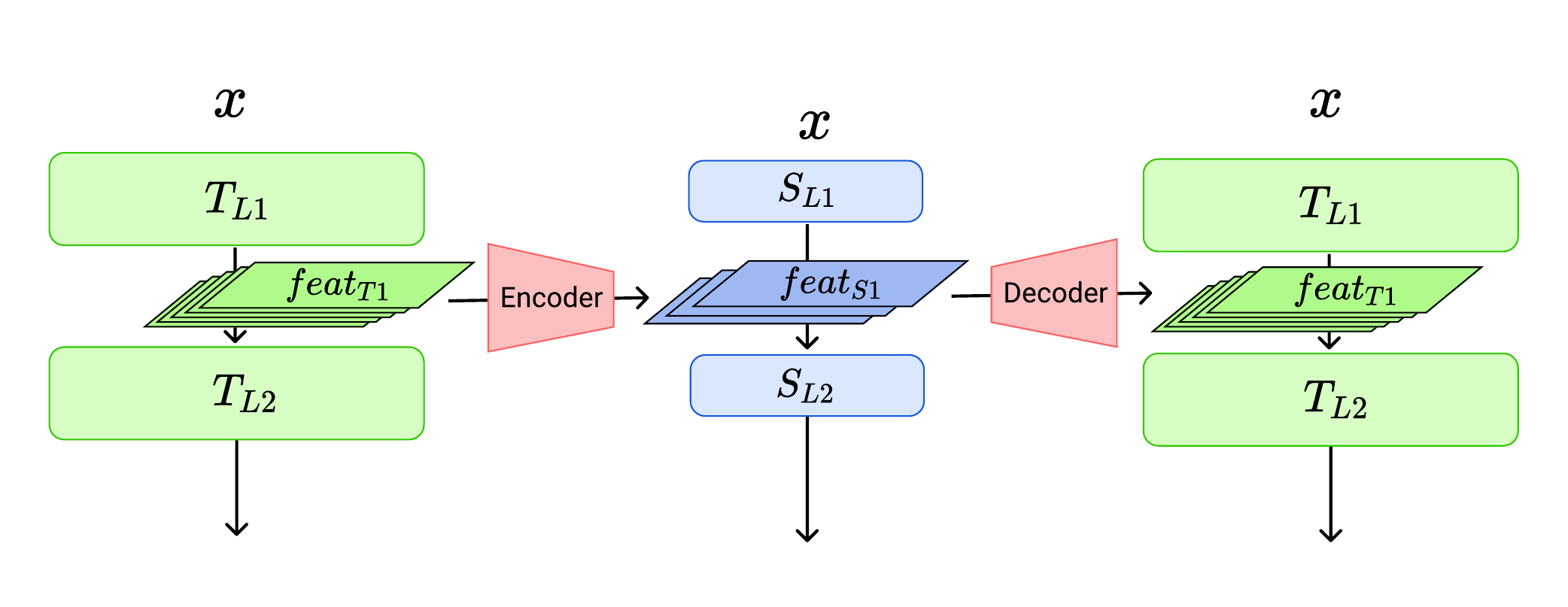}

  \caption{Architecture of the feature converter. Since features from the teacher and student models often differ in dimensionality or channel size, they must be transformed to a common space for effective comparison. Our proposed solution employs a lightweight autoencoder-style converter, consisting of simple linear layers for both the encoder and decoder.}

  \label{fig:feature_converter}

\end{figure}

One of the critical challenges we address is aligning different feature
representations between student and teacher model layers. The goal is to ensure
compatibility between features extracted from corresponding layers, enabling
the progressive loading of teacher model layers into a student model.

We define the features from the student and teacher models as follows
\ref{eq:student_teacher_features}:

\begin{align}
  \label{eq:student_teacher_features}
  \text{feat}_{Si} & = S_{Li} \circ S_{Li-1} \circ \cdots \circ S_{L1}(x),           \\
  \text{feat}_{Ti} & = T_{Li} \circ T_{Li-1} \circ \cdots \circ T_{L1}(x). \nonumber
\end{align}

where $f_{S_{1:i}}$ represents the output feature from layers $1$ to $i$ of the
student model, and $f_{T_{1:i}}$ represents the corresponding feature from the
teacher model.

For conventional CNN based architectures (e.g., VGG, ResNet), these features
typically differ in the number of channels, given by \ref{eq:2}:

\begin{align}
  \label{eq:2}
  \text{feat}_{Si} & \in \mathbb{R}^{C_s \times W \times H},           \\
  \text{feat}_{Ti} & \in \mathbb{R}^{C_t \times W \times H}. \nonumber
\end{align}

where usually $C_s < C_t$.

In transformer-based models such as Vision Transformer (ViT), the features
differ primarily in embedding dimensionality \ref{eq:3}:

\begin{align}
  \label{eq:3}
  \text{feat}_{Si} & \in \mathbb{R}^{\text{num\_blocks}\times d_S},           \\
  \text{feat}_{Ti} & \in \mathbb{R}^{\text{num\_blocks}\times d_T}. \nonumber
\end{align}

where typically $d_s < d_t$.

To convert teacher features to student features, we propose using a lightweight
linear mapping. Specifically, we employ encoder decoder pairs for each layer
\ref{eq:4}, designed to map high dimensional teacher features to low
dimensional student features and vice versa. For CNN based models, these
mappings are implemented as $1 \times 1$ convolutional layers, whereas for
transformer-based models, simple linear layers are used.

Crucially, this feature mapping must be invertible to enable flexible layer
replacement. Suppose we replace a student layer $S_{L2}$ with its teacher
counterpart $T_{L2}$. First, the student feature $\text{feat}_{S1}$ must be
mapped to $\text{feat}_{T1}$ before being processed by $T_{L2}$. Subsequently,
the resulting teacher feature $\text{feat}_{T2}$ needs to be mapped back to
$\text{feat}_{S2}$ to continue processing by the student layer $S_{L3}$. Thus,
bidirectional mappings are essential:

\begin{align}
  \label{eq:4}
  \text{Encoder}_{i}(\text{feat}_{Ti}) & = \text{feat}_{Si},           \\
  \text{Decoder}_{i}(\text{feat}_{Si}) & = \text{feat}_{Ti}. \nonumber
\end{align}

We propose an autoencoder-style\cite{hinton2006reducing} architecture for these
encoder decoder feature converters, parameterized by learnable weights. These
encoders and decoders are jointly optimized with the student model during the
distillation training process, ensuring accurate and efficient feature
alignment for Progressive Weight Loading.

\subsection{Training Strategy for PWL}

\begin{figure}[htbp]

  \centering

  \includegraphics[width=1.0\linewidth]{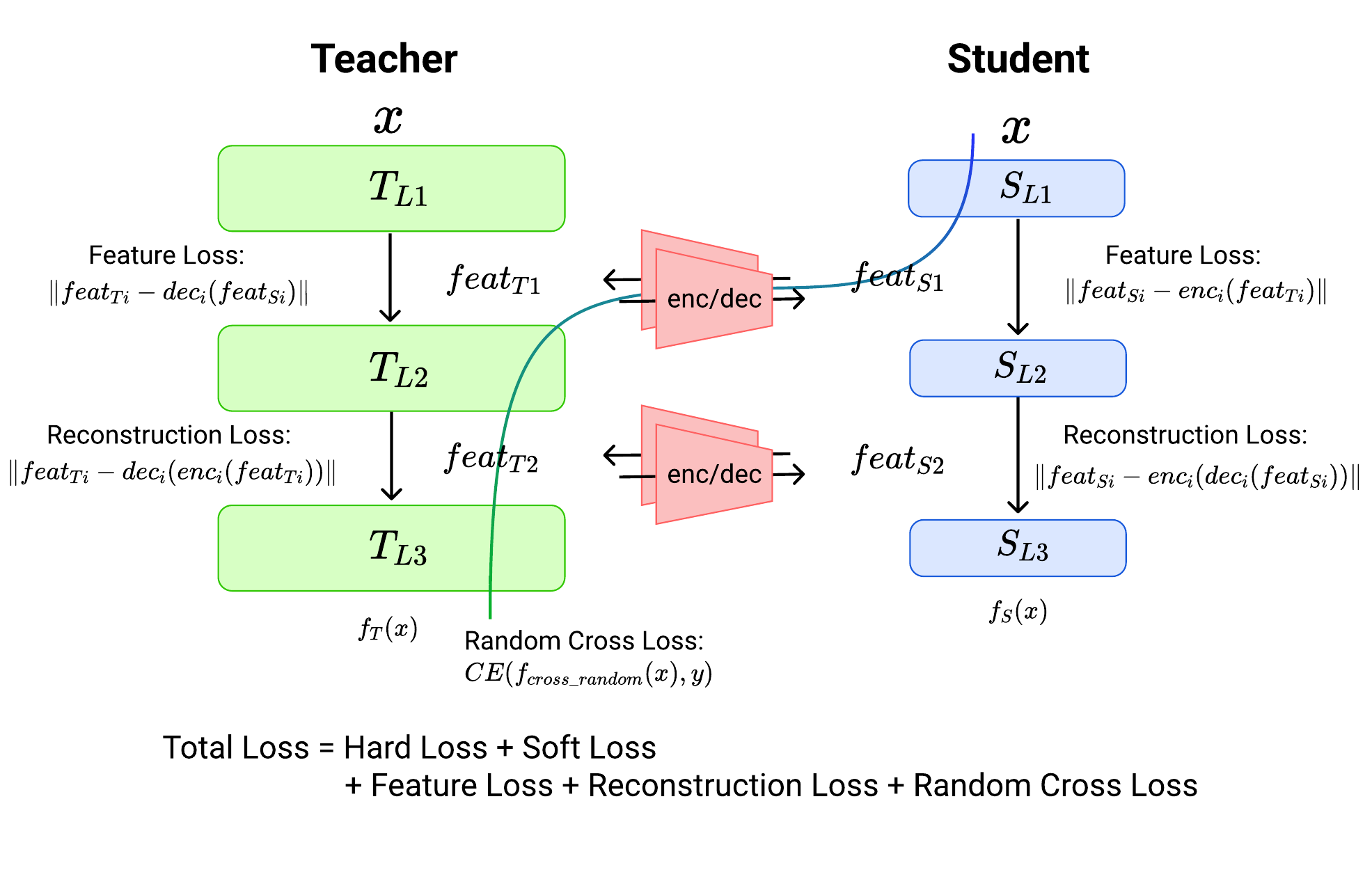}

  \caption{Training strategy for PWL. The student model is optimized using a combination of five losses to support progressive layer replacement. Hard loss and soft loss supervise the student using ground-truth labels and teacher logits, respectively. Feature loss and reconstruction loss encourage alignment of intermediate representations between the student and teacher. Random cross loss mitigates performance degradation when individual student layers are replaced with corresponding teacher layers.}

  \label{fig:pwl_loss}

\end{figure}

The training strategy for PWL involves jointly optimizing the student model and
the encoder decoder\cite{hinton2006reducing} feature converters through a
distillation framework. The Teacher model guides the student model to learn
from its logits and intermediate representations. During this process, the
encoder decoder converters are simultaneously trained to minimize the
discrepancy between corresponding student and teacher features.

Our training objective comprises 4 primary components:

\subsubsection{Distillation Loss} \ref{eq:distill_loss}: The distillation loss \(L_{\text{distill}}\) consists of two components a hard target loss that enforces correct predictions against ground truth labels, and a soft target loss that encourages the student model to mimic the teacher’s output distribution\cite{hinton2015distilling}.

\begin{align}
  L_{\text{distill}} = \alpha \, L_{\text{hard}} + (1 - \alpha) \, L_{\text{soft}}
  \label{eq:distill_loss}
\end{align}

where

\begin{align}
  \label{eq:hard}
  L_{\text{hard}} & = \mathrm{CE}(p_s, y),                                                             \\
  \label{eq:soft}
  L_{\text{soft}} & = T^2 \, \mathrm{KL}\big(\sigma(\frac{z_t}{T}) \,\|\, \sigma(\frac{z_s}{T}) \big).
\end{align}

with \(z_t\) and \(z_s\) being the teacher and student logits respectively,
\(\sigma\) denoting the softmax function, \(y\) the ground truth label, \(T >
1\) the temperature parameter\cite{guo2017calibration} used for softening
logits, and \(\alpha \in [0,1]\) balancing the hard and soft losses.

\subsubsection{Feature Loss} \ref{eq:feature loss}:  This loss ensures feature compatibility across corresponding layers by minimizing the L2 distance between converted features:
\begin{align}
  \label{eq:feature loss}
  L_{feature} = \sum_{i} ||\text{Encoder}_{i}(\text{feat}_{Ti}) - \text{feat}_{Si}||^2 + ||\text{Decoder}_{i}(\text{feat}_{Si}) - \text{feat}_{Si}||^2
\end{align}

\subsubsection{Reconstruction Loss} \ref{eq:reconstruction loss}: We also introduce a reconstruction loss to encourage the encoder-decoder pairs to be near-invertible. This loss
minimizes the discrepancy between the original teacher feature and its reconstruction after passing through
the encoder and decoder:

\begin{multline}
\label{eq:reconstruction loss}
L_{\text{recon}} = \sum_i \left\| \text{feat}_{Ti} 
- \mathrm{Decoder}_i\big(\mathrm{Encoder}_i(\text{feat}_{Ti})\big) \right\|^2 \\
+ \left\| \text{feat}_{Si} 
- \mathrm{Encoder}_i\big(\mathrm{Decoder}_i(\text{feat}_{Si})\big)\right\|^2
\end{multline}

Including this term helps the feature converters preserve essential information
in the teacher features during mapping, ensuring smooth transitions when
progressively replacing student layers with teacher layers. It acts as a
regularizer\cite{vincent2008extracting} preventing the mappings from collapsing
to trivial or lossy transformations and stabilizing training.

\subsubsection{Random Cross Loss} \ref{eq:random cross loss}:
To ensure consistent accuracy regardless of which student layers are replaced by
corresponding teacher layers during progressive loading, we introduce a Random Cross Loss.
In each training iteration, a random subset of student layers is replaced by the corresponding
teacher layers, with encoder and decoder feature converters applied as needed.

Formally, let the full model consist of \(L\) layers, and let \(\mathcal{R}
\subseteq \{1, 2, \ldots, L\}\) denote the random subset of layers replaced by
teacher layers in a given iteration. The mixed model’s output logits, denoted
by \(z_{\text{mix}}\) \ref{eq:z_mix}, are defined as

\begin{align}
  \label{eq:z_mix}
  z_{\text{mix}} = f\big( x; \{ \tilde{L}_i \}_{i=1}^L \big)
\end{align}

where

\begin{align}
  \label{eq:condition of z_mix}
  \tilde{L}_i =
  \begin{cases}
    T_{L_i}(\cdot) & \text{if } i \in \mathcal{R} \text{ (teacher layer with encoder/decoder conversion)}, \\
    S_{L_i}(\cdot) & \text{if } i \notin \mathcal{R} \text{ (student layer)}.
  \end{cases}
\end{align}

Here, \(S_{L_i}\) and \(T_{L_i}\) represent the student and teacher
implementations of the \(i\) th layer, respectively, with feature conversion
applied to teacher layers to ensure compatibility.

The Random Cross Loss is then computed as the cross entropy between the mixed
model's output and the ground truth label \(y\):

\begin{align}
  \label{eq:random cross loss}
  L_{\text{random-cross}} = \mathrm{CE}\big(\sigma(z_{\text{mix}}), y\big)
\end{align}

where \(\sigma(\cdot)\) denotes the softmax function.

This loss encourages the model to maintain stable and high performance across
arbitrary combinations of student and teacher layers during progressive weight
loading.

\subsubsection{Total Loss}  \ref{eq:total_loss}: Combining these losses, the final training objective is:

\begin{align}
  L_{\text{total}} = L_{\text{distill}} + \lambda_1 L_{\text{feature}} + \lambda_2 L_{\text{recon}} + \lambda_3 L_{\text{random-cross}}
  \label{eq:total_loss}
\end{align}

where \(\lambda_1, \lambda_2,\) and \(\lambda_3\) are hyperparameters that
balance the contributions of feature alignment loss, reconstruction loss, and
Random Cross Loss respectively.

\section{Implementation Details}

All source code, including model definitions, training scripts, and evaluation
utilities, is publicly available at:
\url{https://github.com/5yearsKim/ProgressiveWeightLoading}.

\subsection{Framework and Environment}

All models and training routines are implemented in \textbf{PyTorch 2.7} using
\textbf{Python 3.12}. Experiments are conducted on a system equipped with an
\textbf{NVIDIA A100 GPU} and \textbf{CUDA 12.6}.

Model statistics, such as parameter counts and memory usage, are obtained using
the \texttt{torchinfo} library.

\subsection{Dataset and Preprocessing}

We evaluate our method on two benchmark datasets: \textbf{CIFAR-10} and
\textbf{CIFAR-100}\cite{krizhevsky2009learning}. For training, we apply
standard data augmentation techniques commonly used for these
datasets\cite{krizhevsky2012imagenet,szegedy2015going}, including:

\begin{itemize}
  \item \texttt{RandomCrop(32, padding=4)}
  \item \texttt{RandomHorizontalFlip(p=0.5)}
  \item \texttt{RandomRotation(15°)}
\end{itemize}

All images are normalized\cite{krizhevsky2012imagenet} using dataset-specific
mean and standard deviation values prior to input into the model.

\subsection{Model Architecture}

We designed distinct architectures for teacher and student models across three
standard architectures VGG\cite{simonyan2014very}, ResNet\cite{he2016deep}, and
Vision Transformer (ViT)\cite{dosovitskiy2020image}—to rigorously evaluate
Progressive Weight Loading (PWL). Each architecture is divided into four main
blocks, allowing clear comparisons of intermediate representations and
facilitating layer-wise progressive loading.

\subsubsection{VGG Architecture}

For the VGG model, the teacher follows the classical VGG16 style
architecture\cite{simonyan2014very} comprising four convolutional blocks
followed by three fully connected (FC) layers. The student architecture
significantly reduces complexity, adopting a VGG6 style model featuring only
five convolutional layers distributed across four blocks and a single FC layer.
Additionally, each convolutional block's channel size is halved compared to the
teacher, except for the initial block.

\subsubsection{ResNet Architecture}

The ResNet teacher model adopts the widely recognized ResNet-50
architecture\cite{he2016deep}. It consists of four convolutional blocks
containing 3, 4, 6, and 3 residual layers, respectively. Each residual layer
includes two $1 \times 1$ convolutional layers and one $3 \times 3$
convolutional layer, with channel dimensions selected as 64, 128, 256, and 512.
Conversely, the student architecture simplifies to a ResNet-18 style model,
featuring only a single residual layer per block with channel dimensions
reduced by half, except for the first block.

\subsubsection{ViT Architecture}

For the Vision Transformer (ViT), the teacher model uses a ViT-tiny
configuration\cite{dosovitskiy2020image} with 12 transformer layers evenly
divided into four blocks (three layers per block), each with an embedding
dimension of 192. The student variant simplifies this by allocating just one
transformer layer per block, maintaining the same embedding dimension to
facilitate feature mapping between corresponding layers.

Table \ref{tab:model_architectures} summarizes the specific architectural
differences between teacher and student models, highlighting parameter
reductions and resulting memory efficiencies for each architecture.

\begin{table}[htbp]
  \centering
  \caption{Comparison of teacher and student architectures for VGG, ResNet, and ViT. Each model is divided into four corresponding blocks of layers for structural alignment and progressive replacement.}

  \label{tab:model_architectures}

  \begin{tabularx}{\linewidth}{l C C}

    \toprule

    \textbf{Block} & \textbf{Teacher (VGG16 style)}        & \textbf{Student (VGG6 style)} \\

    \midrule

    Block 1        &

    \shortstack[c]{2 Conv $3\times3$ (64)                                                  \\ $\to$ 2 Conv $3\times3$ (128)}

                   & \shortstack[c]{1 Conv $3\times3$ (64)                                 \\ $\to$ 1 Conv $3\times3$(128)} \\

    Block 2        & 3 Conv $3\times3$ (256)               & 1 Conv $3\times3$ (128)       \\

    Block 3        & 3 Conv $3\times3$ (512)               & 1 Conv $3\times3$ (256)       \\

    Block 4        & 3 Conv $3\times3$ (512)               & 1 Conv $3\times3$ (256)       \\

    Post blocks    & 3 FC                                  & 1 FC                          \\

    \midrule

    \#Params       & 14.7M                                 & 0.9M                          \\

    Memory (MB)    & 64.1                                  & 5.1 (8.3\%)                   \\

    \bottomrule

  \end{tabularx}

  \vspace{1em}

  \begin{tabularx}{\linewidth}{l C C}

    \toprule

    \textbf{Block} & \textbf{Teacher (ResNet-50 style)} & \textbf{Student (ResNet-18 style)} \\

    \midrule

    Block 1        & 3 ResNet layers (64)               & 1 ResNet layer (32)                \\

    Block 2        & 4 ResNet layers (128)              & 1 ResNet layer (64)                \\

    Block 3        & 6 ResNet layers (256)              & 1 ResNet layer (128)               \\

    Block 4        & 3 ResNet layers (512)              & 1 ResNet layer (256)               \\

    Post blocks    & Global Pooling + FC                & Global Pooling + FC                \\

    \midrule

    \#Params       & 10.6M                              & 1.2M                               \\

    Memory (MB)    & 61.2                               & 8.9 (14.5\%)                       \\

    \bottomrule

  \end{tabularx}

  \vspace{1em}

  \begin{tabularx}{\linewidth}{l C C}

    \toprule

    \textbf{Block} & \textbf{Teacher (ViT-tiny, 12 layers)} & \textbf{Student (ViT-tiny, 4 layers)} \\

    \midrule

    Block 1        & 3 ViT layers (192 dim)                 & 1 ViT layer (192 dim)                 \\

    Block 2        & 3 ViT layers (192 dim)                 & 1 ViT layer (192 dim)                 \\

    Block 3        & 3 ViT layers (192 dim)                 & 1 ViT layer (192 dim)                 \\

    Block 4        & 3 ViT layers (192 dim)                 & 1 ViT layer (192 dim)                 \\

    Post blocks    & FC                                     & FC                                    \\

    \midrule

    \#Params       & 5.5M                                   & 2.1M                                  \\

    Memory (MB)    & 62.8                                   & 22.7 (36.1\%)                         \\

    \bottomrule
  \end{tabularx}

\end{table}

\subsection{Experiment Setup}

We adopt distinct training strategies for CNN based model(VGG and ResNet) and
for transformer-based models(ViT), to reflect their architectural differences.

For CNN based distillation, we train the student models from scratch using
standard supervised learning with an extended training schedule. Models are
trained for 160 epochs on CIFAR-10 and CIFAR-100 using a cosine annealing
learning rate schedule starting from 5e-2 and decaying to 1e-5. We use the SGD
optimizer with a momentum of 0.9 and weight decay of 5e-4. This prolonged
training stabilizes convergence and facilitates effective feature alignment
with the teacher model.

For transformer-based architectures, we adopt a two stage training process.
First, the student model is pretrained using intermediate representation (IR)
matching and feature loss \ref{eq:feature loss}. Then, the student is finetuned
on CIFAR-10 or CIFAR-100 for 5 epochs using the AdamW optimizer with a fixed
learning rate of 5e-5.

To enhance stability, we assign a smaller learning rate 1/10th of the base rate
to the encoder and decoder layers used for feature conversion. This prevents
overfitting of the converters and ensures consistent performance when student
layers are progressively replaced by teacher layers.

For the loss formulation, we set the balancing coefficient $\alpha$ in distill
loss ($L_{\text{distill}}$) \ref{eq:distill_loss} to 0.6, assigning a weight of
0.6 to the hard target loss ($L_{\text{hard}}$) \ref{eq:hard} and 0.4 to the
soft target loss ($L_{\text{soft}}$) \ref{eq:soft}. For the total loss
\ref{eq:total_loss}, we configure the loss weights as follows: $\lambda_1 =
  1.0$ and $\lambda_2 = 1.0$ for the feature loss ($L_{\text{feature}}$)
\ref{eq:feature loss} and reconstruction loss ($L_{\text{recon}}$)
\ref{eq:reconstruction loss}, respectively. The weight for the Random Cross
Loss ($L_{\text{random-cross}}$) \ref{eq:random cross loss} is set to
$\lambda_3 = 1.8$, reflecting its greater influence in stabilizing performance
during progressive layer replacement.

\section{Result and Analysis}

\begin{table}[ht]

  \centering
  \caption{Comparison of distillation performance with and without PWL training. The student model trained with PWL shows no degradation in performance compared to standard distillation.}

  \label{tab:distillation}
  \begin{tabularx}{\textwidth}{l *{3}{C}|| *{3}{C}}

    \toprule
                                       &
    \multicolumn{3}{c||}{CIFAR10}
                                       &
    \multicolumn{3}{c}{CIFAR100}                                                                                                       \\

    \cmidrule(lr){2-4} \cmidrule(lr){5-7}

    Model                              & VGG           & ResNet        & ViT           & VGG           & ResNet        & ViT           \\

    \midrule

    Teacher                            & 93.8          & 94.8          & 97.4          & 74.2          & 75.7          & 82.3          \\

    Student (w/o PWL training)         & 91.3          & 92.0          & 94.3          & 70.8          & 72.4          & 75.2          \\

    \textbf{Student (w/ PWL training)} & \textbf{91.7} & \textbf{92.9} & \textbf{94.1} & \textbf{71.1} & \textbf{72.1} & \textbf{74.6} \\

    \bottomrule

  \end{tabularx}

\end{table}

We evaluate the classification accuracy of student models trained with and
without the proposed PWL loss. As shown in Table~\ref{tab:distillation}, the
inclusion of PWL loss leads to negligible or minor performance changes across
architectures. Specifically, VGG and ResNet show marginal improvements, while
ViT experiences a slight decrease in accuracy. These results confirm that
incorporating PWL training does not negatively impact the overall effectiveness
of knowledge distillation. Importantly, it enables the capability of
progressively loading teacher weights into the student model without degrading
performance.

\begin{table}[ht]

  \centering

  \caption{Gradual performance improvements as teacher weights are progressively
    loaded from input to output layers in VGG, ResNet, and ViT models.}

  \label{tab:gradual_improvement}

  \begin{tabularx}{\textwidth}{l *{2}{C} C}

    \toprule

    Layers Loaded                          & CIFAR10 (\%) & CIFAR100 (\%) & Memory Loaded(MB) \\

    \midrule


    \addlinespace

    \multicolumn{4}{c}{\bfseries VGGNet}                                                      \\

    \addlinespace

    Student ($S_{L1}S_{L2}S_{L3}S_{L4}$)   & 91.7         & 71.1          & 5.1               \\

    $T_{L1}\to S_{L2}\to S_{L3}\to S_{L4}$ & 92.1         & 70.7          & 8.3               \\

    $T_{L1}\to T_{L2}\to S_{L3}\to S_{L4}$ & 93.1         & 72.5          & 16.3              \\

    $T_{L1}\to T_{L2}\to T_{L3}\to S_{L4}$ & 92.7         & 73.9          & 39.3              \\

    Teacher ($T_{L1}T_{L2}T_{L3}T_{L4}$)   & 93.8         & 74.2          & 64.1              \\

    \addlinespace


    \addlinespace

    \cmidrule[0.75pt]{1-4}

    \multicolumn{4}{c}{\bfseries ResNet}                                                      \\

    \addlinespace

    Student ($S_{L1}S_{L2}S_{L3}S_{L4}$)   & 92.3         & 71.2          & 8.9               \\

    $S_{L1}\to S_{L2}\to S_{L3}\to T_{L4}$ & 91.9         & 71.3          & 12.9              \\

    $S_{L1}\to S_{L2}\to T_{L3}\to T_{L4}$ & 92.9         & 73.0          & 26.6              \\

    $S_{L1}\to T_{L2}\to T_{L3}\to T_{L4}$ & 94.4         & 72.5          & 42.1              \\

    Teacher ($T_{L1}T_{L2}T_{L3}T_{L4}$)   & 94.8         & 73.4          & 61.2              \\

    \addlinespace


    \addlinespace

    \cmidrule[0.75pt]{1-4}

    \multicolumn{4}{c}{\bfseries ViT}                                                         \\

    \addlinespace

    Student ($S_{L1}S_{L2}S_{L3}S_{L4}$)   & 94.3         & 74.6          & 22.7              \\

    $S_{L1}\to S_{L2}\to S_{L3}\to S_{L4}$ & 95.5         & 76.2          & 32.2              \\

    $S_{L1}\to S_{L2}\to S_{L3}\to S_{L4}$ & 96.4         & 79.8          & 43.3              \\

    $S_{L1}\to S_{L2}\to S_{L3}\to T_{L4}$ & 97.1         & 81.4          & 52.5              \\

    Teacher ($T_{L1}T_{L2}T_{L3}T_{L4}$)   & 97.8         & 82.3          & 62.8              \\

    \bottomrule

  \end{tabularx}

\end{table}

Table~\ref{tab:gradual_improvement} presents the classification performance and
memory usage as teacher layers are progressively loaded into the student model,
block by block, from input to output. For all three architectures—VGG, ResNet,
and ViT—we observe a clear upward trend in accuracy on both CIFAR-10 and
CIFAR-100 datasets as more teacher layers are introduced. This confirms the
effectiveness of Progressive Weight Loading (PWL) in bridging the performance
gap between the student and teacher models.

Specifically, for VGG and ResNet, accuracy improves steadily while memory usage
increases in proportion to the number of loaded teacher blocks. ViT exhibits a
more pronounced performance gain, particularly on CIFAR-100, indicating that
transformer-based architectures may benefit more from incremental upgrades.
Notably, this progressive transition does not result in abrupt changes in
performance, demonstrating the stability and smooth integration of teacher
layers at each stage.

These results highlight that PWL enables flexible trade-offs between accuracy
and resource usage, allowing systems to scale performance in response to
available memory or computational constraints.



































\begin{figure}[htbp]
  \centering
  \includegraphics[width=\linewidth]{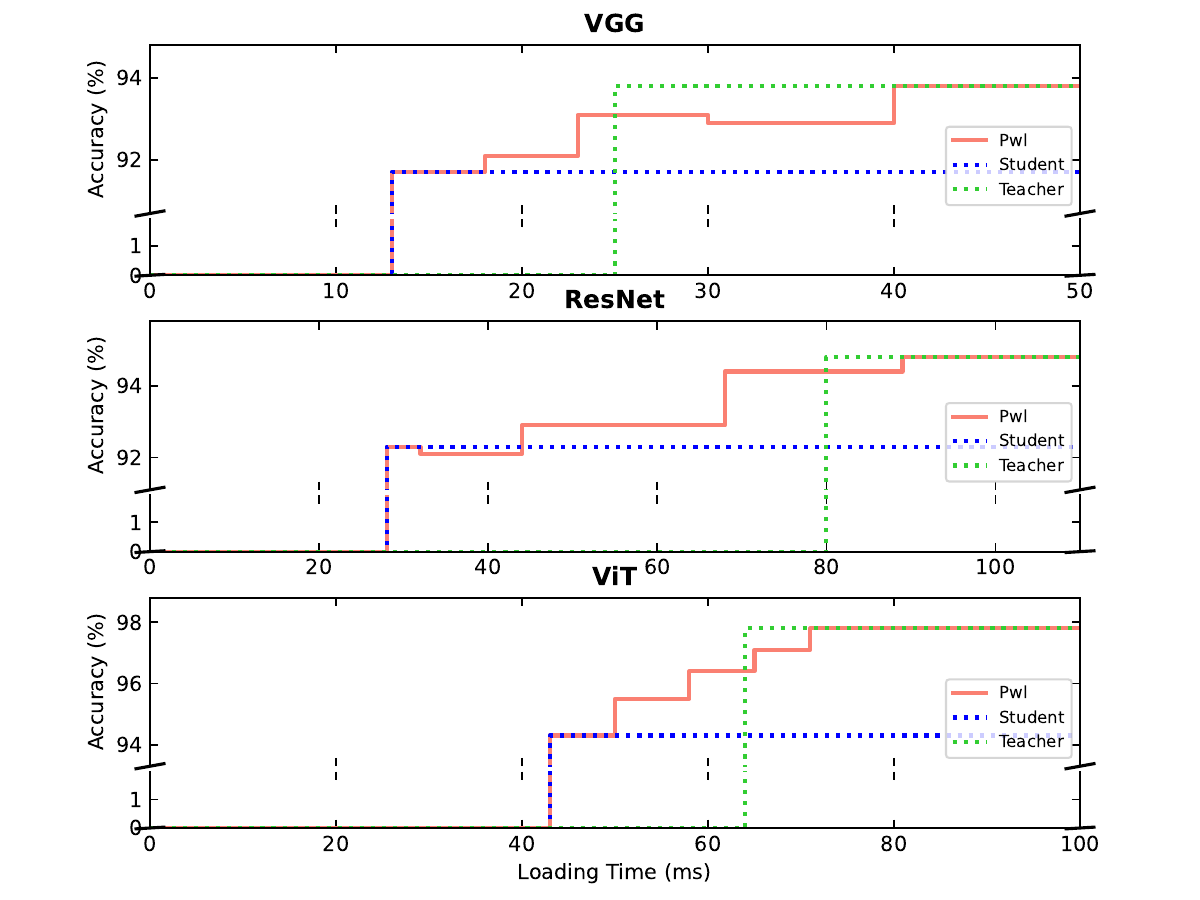}
  \caption{Loading time vs.\ CIFAR-10 accuracy for student, PWL stages, and teacher models across VGG, ResNet, and ViT architectures. PWL delivers inference speed on par with the student model while steadily increasing accuracy as layers are replaced with teacher weights.}
  \label{fig:pwl_loading_time_detail}
\end{figure}

We evaluate the loading time of the model for VGG, ResNet and ViT. As shown in
figure~\ref{fig:pwl_loading_time_detail}, because PWL first loads the entire
student model, the time to first inference is identical to the total time
required to load the student model itself. We then incrementally replaced each
student layer with its teacher counterpart, while still allowing the unreplaced
student layers to perform inference, and measured the time for each step. The
total loading time of teacher layers with PWL exceeds the total loading time of
the student model, but it has a benefit in that it can still inference with
gradually improving performance during the loading time.

Figure~\ref{fig:pwl_loading_time_detail} presents a comparison of loading times
for the student model, the full teacher model, and the progressive layer by
layer loading process using PWL across VGG, ResNet, and ViT architectures. The
"Student Total" row reflects the initial loading time required to load the
entire student model, while the "Teacher Total" row shows the time needed to
load the complete teacher model from scratch.

In the "Student $\to$ Teacher Replace" row, we measure the cumulative time
taken as teacher layers are incrementally loaded on top of the already
initialized student model. Although the total time for progressive loading
exceeds the teacher model's loading time, PWL offers a critical advantage:
inference can begin immediately with the student model and continue throughout
the loading process with gradually improving performance. This incremental and
non blocking behavior is especially valuable for latency sensitive scenarios,
where maintaining continuous availability is more important than waiting for a
full teacher model to load.

\begin{table}[ht]

  \centering
  \caption{System requirements for ResNet on CIFAR-10. PWL achieves fast initial inference comparable to the student model, while progressively reaching the maximum accuracy of the teacher model.}

  \label{tab:resnet_system_req}

  \begin{tabular}{lccc}

    \toprule

            & Initial Inference Time       & Maximum Accuracy & Memory Requirement \\

    \midrule

    Student & \textbf{24.12 ms}            & 92.9\%           & 8.9 MB             \\

    Teacher & 65.36 ms                     & \textbf{94.8\%}  & 61.2 MB            \\

    \textbf{PWL}

            & \textbf{24.3 ms}

            & \textbf{94.8\%}

            & \textbf{8 – 62MB (flexible)}                                         \\

    \bottomrule

  \end{tabular}

\end{table}

Table~\ref{tab:resnet_system_req} summarizes the practical benefits of the
proposed PWL method. For ResNet with CIFAR-10 dataset, PWL achieves the same
maximum accuracy (94.8\%) but with a significantly reduced initial inference
time—comparable to that of the lightweight student model, compared to teacher
model. Moreover, while the teacher model demands a fixed memory footprint of
61.2\,MB, PWL offers a flexible memory range (8-62\,MB), allowing it to adapt
to system constraints by progressively loading teacher layers. This
adaptability enables efficient resource utilization and fast initial responses,
making PWL particularly suitable for deployment in latency sensitive or memory
constrained environments.

\section{Ablation Study}

Table~\ref{tab:loading_order} presents an ablation study evaluating the impact
of different teacher layer loading orders on classification accuracy for VGG,
ResNet, and ViT across CIFAR-10 and CIFAR-100. We tested three strategies:
\textbf{prefix loading} (replacing layers from input to output), \textbf{suffix
  loading} (from output to input), and \textbf{contiguous block loading}
(replacing a block of intermediate layers).

For VGG and ResNet, the accuracy was relatively stable regardless of loading
order, though prefix loading consistently yielded slightly better results. In
contrast, ViT exhibited a significant sensitivity to the loading strategy.
Prefix loading led to smooth and progressive accuracy gains, whereas suffix and
contiguous loading resulted in performance degradation.

These results highlight that for PWL to be effective especially in attention
based architectures like ViT—progressive loading from early layers (prefix
loading) is the most robust and optimal strategy.

\begin{table}[ht]
  \centering
  \caption{Ablation study on the effectiveness of layer replacement order, conducted on VGG, ResNet, and ViT using CIFAR-10 and CIFAR-100. Results show that replacing layers in prefix loading—from input to output—yields the most stable and consistent performance improvement.}

  \label{tab:loading_order}
  \begin{tabularx}{\textwidth}{l *{6}{C}}

    \toprule

    Loading Order                          &

    \multicolumn{3}{c}{CIFAR10}            &

    \multicolumn{3}{c}{CIFAR100}                                                         \\

    \cmidrule(lr){2-4} \cmidrule(lr){5-7}

                                           & VGG  & ResNet & ViT  & VGG  & ResNet & ViT  \\

    \midrule
    \addlinespace
    \multicolumn{7}{c}{\bfseries Prefix Loading}                                         \\
    \addlinespace
    Student($S_{L1}S_{L2}S_{L3}S_{L4}$)    & 91.7 & 92.3   & 94.3 & 71.1 & 72.1   & 74.6 \\
    $T_{L1}\to S_{L2}\to S_{L3}\to S_{L4}$ & 92.1 & 91.9   & 95.5 & 70.6 & 71.4   & 76.2 \\
    $T_{L1}\to T_{L2}\to S_{L3}\to S_{L4}$ & 93.1 & 92.9   & 96.4 & 72.5 & 72.8   & 79.8 \\
    $T_{L1}\to T_{L2}\to T_{L3}\to S_{L4}$ & 92.7 & 94.4   & 97.1 & 73.8 & 73.6   & 81.4 \\
    Teacher($T_{L1}T_{L2}T_{L3}T_{L4}$)    & 93.8 & 94.8   & 97.4 & 74.2 & 75.7   & 82.3 \\
    \addlinespace
    \cmidrule[0.75pt]{1-7}
    \addlinespace
    \multicolumn{7}{c}{\bfseries Suffix Loading}                                         \\
    \addlinespace
    Student($S_{L1}S_{L2}S_{L3}S_{L4}$)    & 91.7 & 92.3   & 94.3 & 71.1 & 72.1   & 74.6 \\
    $S_{L1}\to S_{L2}\to S_{L3}\to T_{L4}$ & 91.0 & 88.8   & 82.1 & 65.6 & 69.2   & 61.3 \\
    $S_{L1}\to S_{L2}\to T_{L3}\to T_{L4}$ & 90.3 & 93.4   & 84.2 & 67.6 & 73.7   & 64.7 \\
    $S_{L1}\to T_{L2}\to T_{L3}\to T_{L4}$ & 91.4 & 94.3   & 87.5 & 69.1 & 74.5   & 68.8 \\
    Teacher($T_{L1}T_{L2}T_{L3}T_{L4}$)    & 93.8 & 94.8   & 97.4 & 74.2 & 75.7   & 82.3 \\
    \addlinespace
    \cmidrule[0.75pt]{1-7}
    \addlinespace
    \multicolumn{7}{c}{\bfseries Contiguous Block Loading}                               \\
    \addlinespace
    Student($S_{L1}S_{L2}S_{L3}S_{L4}$)    & 91.7 & 91.8   & 94.3 & 53.1 & 72.1   & 74.6 \\
    $S_{L1}\to T_{L2}\to S_{L3}\to S_{L4}$ & 86.1 & 88.1   & 84.5 & 59.4 & 65.2   & 62.1 \\
    $S_{L1}\to S_{L2}\to T_{L3}\to S_{L4}$ & 88.4 & 90.2   & 81.2 & 64.1 & 68.8   & 59.7 \\
    $S_{L1}\to T_{L2}\to T_{L3}\to S_{L4}$ & 89.8 & 89.1   & 78.2 & 65.8 & 71.0   & 58.3 \\
    Teacher($T_{L1}T_{L2}T_{L3}T_{L4}$)    & 93.8 & 94.2   & 97.4 & 56.1 & 75.7   & 82.3 \\
    \bottomrule
  \end{tabularx}
\end{table}

\begin{figure}[htbp]
  \centering
  \includegraphics[width=\linewidth]{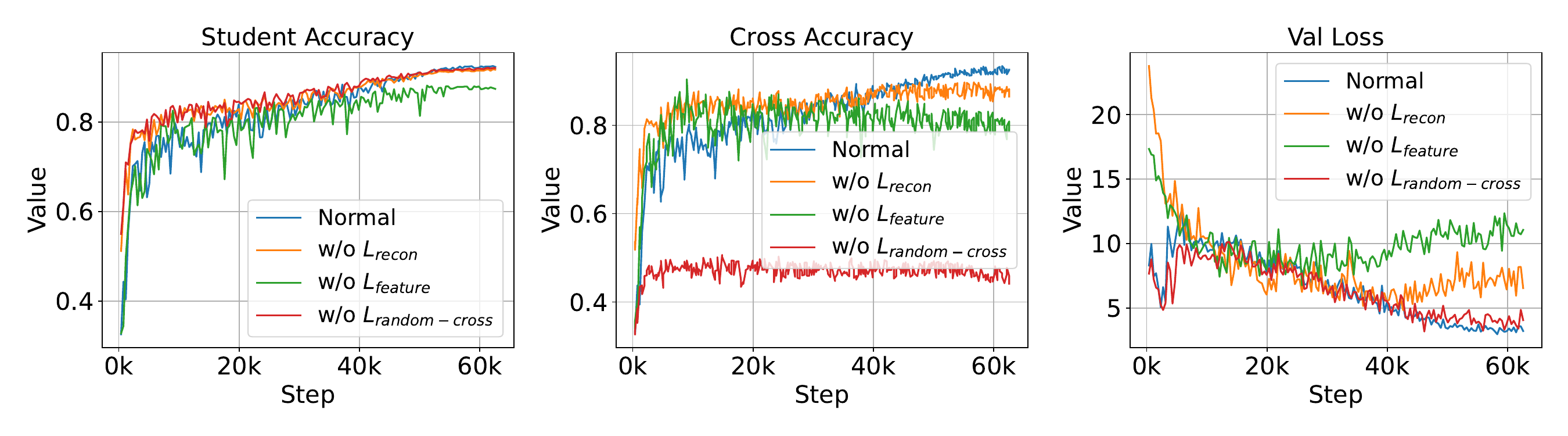}
  \caption{Training metrics—student accuracy, cross accuracy, and validation loss—for ResNet on CIFAR-10, illustrating the impact of ablating feature loss, reconstruction loss, and random-cross loss.}
  \label{fig:ablation_loss}
\end{figure}

Also, we assessed the individual effectiveness of each component in the
proposed PWL training. Table~\ref{tab:loss_effectiveness} and
figure~\ref{fig:ablation_loss} presents an ablation study evaluating the
contribution of each component in the proposed loss function
(Equation~\ref{eq:total_loss}) on the ResNet model trained with CIFAR-10. Two
evaluation metrics are used: the classification accuracy of the final distilled
student model, and the \textbf{Cross Accuracy}, which is defined as the mean
accuracy across intermediate hybrid models where teacher layers progressively
replace student layers. For example, with three blocks, Cross Accuracy includes
the average of models like $T_{L1}S_{L2}S_{L3}$ and $T_{L1}T_{L2}S_{L3}$.

The results demonstrate that $L_{feature}$ plays a critical role in both
student accuracy and cross accuracy. This indicates that aligning internal
feature representations not only improves final performance but also enhances
model stability during layer-wise replacement. On the other hand, $L_{recon}$
and $L_{random-cross}$ do not significantly affect the distilled student's
final accuracy but greatly improve cross accuracy. Notably, removing
$L_{random-cross}$ leads to a dramatic drop in cross accuracy (from 93.1\% to
44.2\%), confirming its essential role in ensuring seamless replacement between
student and teacher layers during progressive loading. These findings support
that each component of the proposed loss function contributes to different but
complementary aspects of PWL effectiveness.

\begin{table}[ht]

  \centering
  \caption{Ablation study on the effectiveness of each loss component for ResNet on CIFAR-10. We evaluate the impact of feature loss, reconstruction loss, and random cross loss, each of which contributes to improving either the final student accuracy or the cross-layer accuracy during progressive weight replacement.}

  \label{tab:loss_effectiveness}

  \begin{tabular}{lcc}

    \toprule

    Configuration          & Student Acc (\%)      & Cross Acc (mean, \%)  \\

    \midrule

    Normal                 & 92.8                  & 93.1                  \\

    w/o $L_{recon}$        & 92.4                  & 88.9 (\(\downarrow\)) \\

    w/o $L_{feature}$      & 88.4 (\(\downarrow\)) & 82.0 (\(\downarrow\)) \\

    w/o $L_{random-cross}$ & 92.3                  & 44.2 (\(\downarrow\)) \\

    \bottomrule

  \end{tabular}

\end{table}

\section{Discussion and Future Work}

\subsection{Invertibility of the Feature Converter}
An important point of discussion is whether the feature converter needs to be
invertible. We conclude that invertibility is essential when enabling arbitrary
layer replacement between the student and teacher models, rather than following
a fixed, sequential order. Our ablation study table~\ref{tab:loading_order}
shows extensive examples of arbitrary layer replacement order sequence. Also, as
shown in Table~\ref{tab:loss_effectiveness}, the invertibility from
reconstruction loss contributes to cross accuracy, indicating that preserving
information through the converter is beneficial for seamless layer-wise
replacement.

We also examined the possibility of enforcing strict invertibility using QR
decomposition\cite{francis1962qr} or normalizing flow-based
architectures\cite{papamakarios2021normalizing,dinh2016density,kingma2018glow}.
However, to maintain simplicity and computational efficiency, we opted for an
autoencoder-style feature converter.

\subsection{Combining with Other Model Compression Methods}
Model compression techniques such as
quantization\cite{hubara2018quantized,esser2019learned,frantar2022gptq} and
pruning\cite{han2016deepcompressioncompressingdeep,molchanov2016pruning,molchanov2017variational}
are essential for deploying deep learning models in resource-constrained
environments like mobile devices. Progressive Weight Loading (PWL) is
compatible with these techniques and can further enhance efficiency by enabling
faster model loading and earlier inference.

Although integrating PWL with compression methods is beyond the scope of this
study, it presents a promising direction for future research. Such a
combination could lead to highly optimized deployments of deep learning models,
particularly in latency-sensitive or memory-constrained applications.

\subsection{Application to Other Domains}
The need for deploying large-scale models extends beyond vision tasks to
domains such as text-to-speech (TTS), natural language processing (NLP), and
audio processing. In this work, we validate the PWL approach on image
classification tasks as a proof of concept.

Given the effectiveness of PWL on transformer-based models like ViT, we
anticipate its potential applicability to other domains, particularly in NLP
and audio, where large transformer models are prevalent. Exploring these
applications remains an important direction for future work.

\section{Conclusion}

We propose Progressive Weight Loading (PWL), a novel approach that enables
fast initial inference and efficient performance-memory trade-offs by
progressively replacing the student model's layers with those of a pre-trained
teacher model. To support this, we introduced a training method that aligns
intermediate feature representations between student and teacher layers,
ensuring smooth and consistent performance gains as layers are incrementally
upgraded.

Through extensive experiments across VGG, ResNet, and ViT architectures, we
demonstrated that PWL not only maintains distillation performance but also
allows for dynamic and flexible deployment, making it particularly suitable for
latency-sensitive and memory-constrained environments. This study demonstrates
that the PWL technique can be applied to domains where large scale Transformer
models are pervasive such as natural language processing and speech and future
work will focus on validating PWL in these settings and rigorously quantifying
its benefits for model scalability and responsiveness.

\bibliographystyle{splncs04}
\bibliography{references}

\clearpage
\appendix
\section*{Appendix}

\section{Feature Converter Capacity Study}

\begin{table}[ht]
  \centering
  \caption{Different size of feature converters.}
  \label{tab:feature_converter_capacity}
  \begin{tabular}{
      l
      c @{\hspace{1em}}
      c @{\hspace{1em}}
      c @{\hspace{1em}}
    }
    \toprule
     & \textbf{Tiny}                & \textbf{Medium} & \textbf{Heavy} \\
    \midrule
    \textbf{Architecture}
     & \makecell[c]{Single                                             \\ Linear}
     & \makecell[c]{Two‐layer MLP                                      \\with bottleneck}
     & \makecell[c]{Three‐layer MLP                                    \\with non‐linearities}     \\
    \textbf{\#params}
     & 98k
     & 163k
     & 310k                                                            \\
    \bottomrule
  \end{tabular}
\end{table}

\begin{figure}[htbp]
  \centering
  \includegraphics[width=\linewidth]{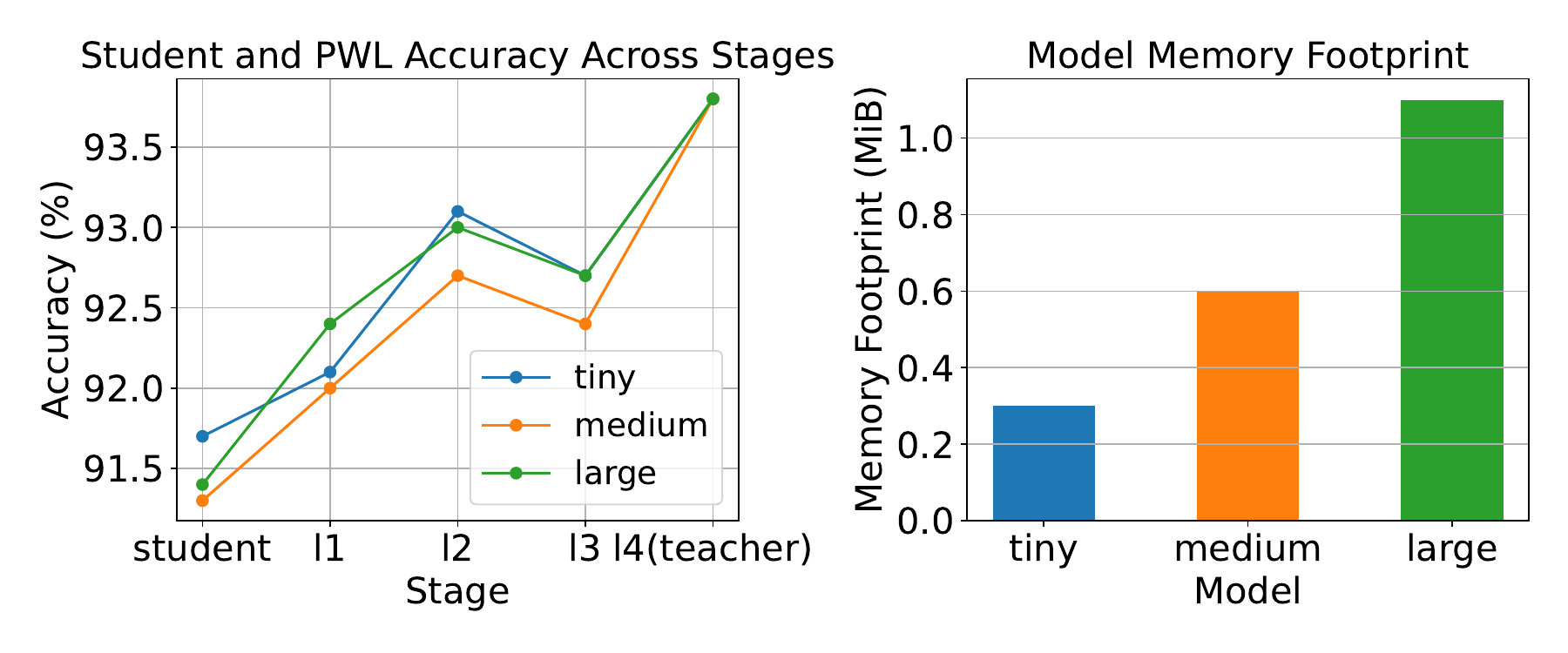}
  \caption{Effect of feature converter capacity on ResNet accuracy for CIFAR-10.}
  \label{fig:feature_converter_capacity}
\end{figure}

We evaluate three feature converter architectures—Tiny, Medium, and Heavy—to
understand how converter capacity affects both inference accuracy and memory
footprint during progressive weight loading.
\begin{itemize}
  \item \textbf{Tiny}: a single linear layer (equivalent to a 1x1 convolution).
  \item \textbf{Medium}: a two-layer MLP with a bottleneck.
  \item \textbf{Heavy}: a three-layer MLP with nonlinear activations.
\end{itemize}
We integrate each converter into a ResNet backbone trained on CIFAR-10 and measure (1) the student accuracy after each layer replacement (cross accuracy), and (2) the converter’s parameter count. Table~\ref{tab:feature_converter_capacity} and Figure~\ref{fig:feature_converter_capacity} summarize these results.

Despite a more than threefold increase in parameters—from 98 k (Tiny) to 310 k
(Heavy)—all three converters achieve nearly identical final accuracy (within
0.3 \%). This negligible performance gain, combined with the stricter memory
constraints of mobile applications, motivates our choice of the Tiny converter
for all subsequent experiments.

\section{Applying PWL to Extremely Small Student Model}

\begin{table}[ht]

  \centering

  \caption{Applying PWL to a highly compact student network. LeNet-5 student model is replaced to VGG16 teacher model}

  \label{tab:lenet_to_vgg}

  \begin{tabularx}{\textwidth}{l *{2}{C} C}

    \toprule

    Layers Loaded                          & CIFAR10 (\%) & CIFAR100 (\%) & Memory Loaded(MiB) \\

    \midrule


    \addlinespace

    \multicolumn{4}{c}{\bfseries LeNet5 -> VGGNet}                                             \\

    \addlinespace

    Student ($S_{L1}S_{L2}S_{L3}S_{L4}$)   & 68.4         & 54.9          & <1MiB              \\

    $T_{L1}\to S_{L2}\to S_{L3}\to S_{L4}$ & 76.1         & 58.7          & 6.3                \\

    $T_{L1}\to T_{L2}\to S_{L3}\to S_{L4}$ & 78.0         & 62.1          & 12.1               \\

    $T_{L1}\to T_{L2}\to T_{L3}\to S_{L4}$ & 82.1         & 65.7          & 31.8               \\

    Teacher ($T_{L1}T_{L2}T_{L3}T_{L4}$)   & 93.8         & 74.2          & 64.1               \\

    \addlinespace

    \bottomrule

  \end{tabularx}
\end{table}

We investigated whether Progressive Weight Loading (PWL) can be effectively
applied to extremely compact student models, particularly for scenarios where
ultra-fast initial inference is prioritized over model accuracy—such as in
real-time operating systems (RTOS) or resource-constrained embedded
environments.

To this end, we adopted LeNet-5\cite{lecun2002gradient} as the student model
due to its minimal size (less than 1MiB) and experimented with distilling
knowledge from a VGG16 teacher model. Our goal was to evaluate whether PWL
could progressively enhance the student model's performance by incrementally
replacing its layers with those of the teacher.

As shown in Table~\ref{tab:lenet_to_vgg}, we observe a clear performance
improvement on both CIFAR-10 and CIFAR-100 as more layers are replaced.
Starting from a baseline LeNet-5, performance improves steadily with each layer
swapped, culminating in the full teacher model. This result highlights the
potential of PWL even for extremely lightweight models, making it a promising
strategy for systems where memory is scarce and rapid model activation is
critical.

\section{Attention Map for ViT}

\begin{figure}[htbp]
  \centering
  \includegraphics[width=\linewidth]{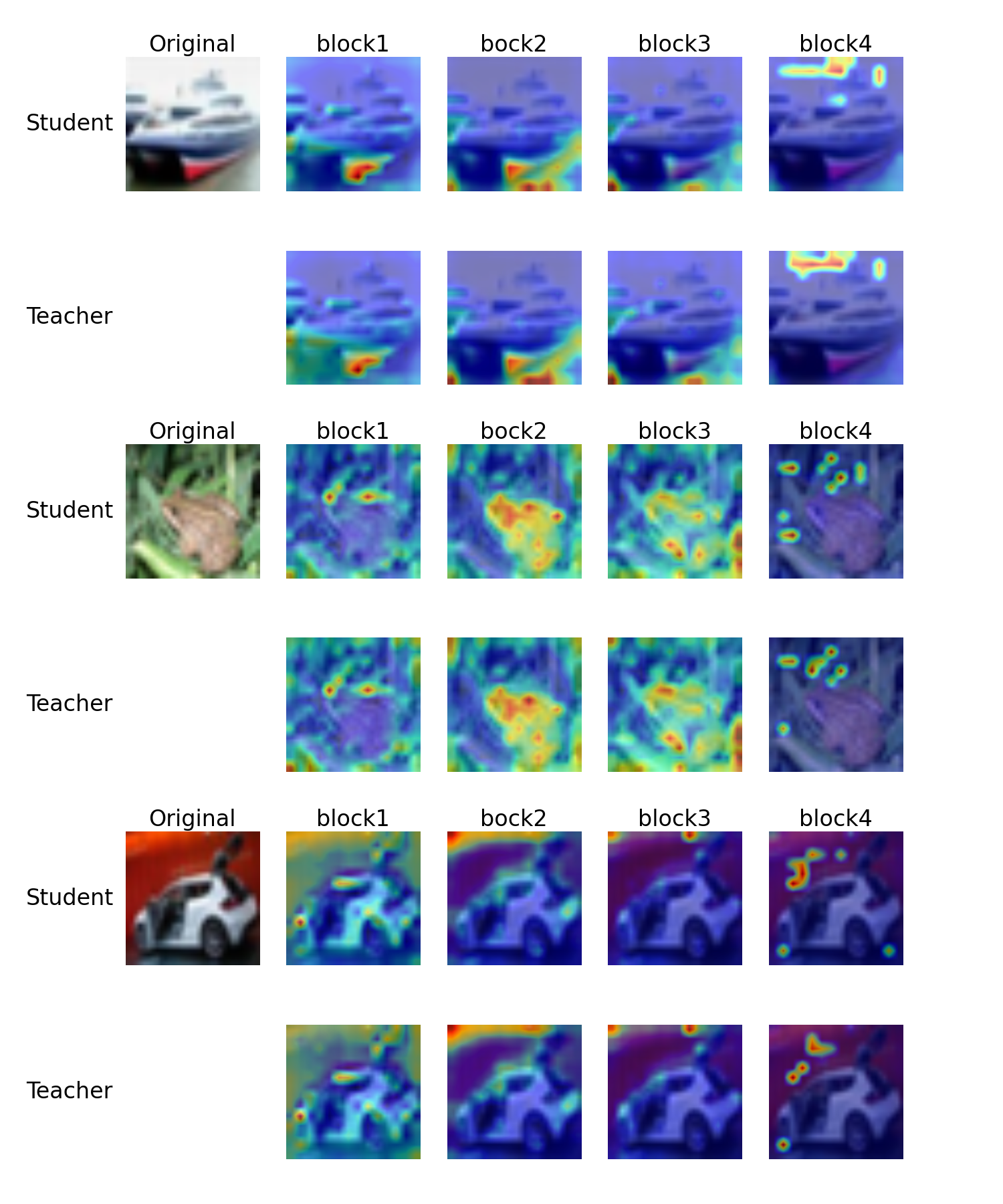}
  \caption{Comparison of attention maps from the final layer of each block for both student and teacher models.}
  \label{fig:attention_map}
\end{figure}

Figure~\ref{fig:attention_map} presents the class-token attention overlaid on a
CIFAR-10 image for both the student (with PWL) and teacher ViT models. For each
transformer block, we extract the attention weights from the block's final
layer and upsample them to the input resolution. The heat-maps reveal that, at
every block boundary, the student model attends to essentially the similar
image regions as the teacher. This strong correspondence in spatial focus
demonstrates that PWL not only maintains predictive accuracy but also preserves
the teacher's interpretability and feature-localization behavior throughout
progressive weight replacement.

\end{document}